%% file: main.tex
\documentclass[conference]{IEEEtran}
\IEEEoverridecommandlockouts
\usepackage{cite}
\usepackage{amsmath,amssymb,amsfonts}
\usepackage{algorithmic}
\usepackage{graphicx}
\usepackage{textcomp}

\usepackage{adjustbox}

\usepackage{multirow}

\usepackage[table]{xcolor}
\usepackage{algorithmic}
\usepackage{amsthm}
\usepackage{booktabs}
\usepackage{algorithm}
\usepackage{bm}

\usepackage{subcaption} 

\def\BibTeX{{\rm B\kern-.05em{\sc i\kern-.025em b}\kern-.08em
    T\kern-.1667em\lower.7ex\hbox{E}\kern-.125emX}}
\begin{document}

\title{TrojFlow: Flow Models are Natural Targets for Trojan Attacks}

\author{\IEEEauthorblockN{Zhengyang Qi}
\IEEEauthorblockA{\textit{University of Science and } \\
\textit{Technology of China}\\
Hefei, China \\
miloq@mail.ustc.edu.cn}
\and
\IEEEauthorblockN{Xiaohua Xu* \thanks{*Corresponding author.}}
\IEEEauthorblockA{\textit{University of Science and } \\
\textit{Technology of China}\\
Hefei, China \\
xiaohuaxu@ustc.edu.cn}
}

\maketitle
\input{0_abstract}
\input{1_introduction}

\input{3_preliminaries}
\input{4_method}

\input{5_experiment}

\input{6_Concolusion}

\bibliographystyle{IEEEbib}
\bibliography{main/main}

\end{document}

%% file: 0_abstract.tex
\begin{abstract}
Flow-based generative models (FMs) have rapidly advanced as a method for mapping noise to data, its efficient training and sampling process makes it widely applicable in various fields. FMs can be viewed as a variant of diffusion models (DMs). At the same time, previous studies have shown that DMs are vulnerable to Trojan/Backdoor attacks, a type of output manipulation attack triggered by a maliciously embedded pattern at model input. We found that Trojan attacks on generative models are essentially equivalent to image transfer tasks from the backdoor distribution to the target distribution, the unique ability of FMs to fit any two arbitrary distributions significantly simplifies the training and sampling setups for attacking FMs, making them inherently natural targets for backdoor attacks. In this paper, we propose TrojFlow, exploring the vulnerabilities of FMs through Trojan attacks. In particular, we consider various attack settings and their combinations and thoroughly explore whether existing defense methods for DMs can effectively defend against our proposed attack scenarios.  We evaluate TrojFlow on CIFAR-10 and CelebA datasets,  our experiments show that our method can compromise FMs with high utility and specificity, and can easily break through existing defense mechanisms.
\end{abstract}


%% file: 1_introduction.tex
\section{Introduction}

Recently, diffusion models (DMs)\cite{ddpm,ddim,scorematching}   demonstrated impressive performance on generative tasks like image synthesis\cite{stablediffusion}, speech synthesis\cite{diffwave} and molecule  generation\cite{EDM,Digress}. However, a disadvantage is that DMs often require hundreds to thousands of time steps to generate high-quality images, making them more computationally expensive compared to other generative models like GAN\cite{gan} or VAE\cite{vae}. Some works focus on accelerating sampling in DMs\cite{ddim,fastsample}, while another group of researchers rethought the optimization objectives of DMs and proposed flow-based generative models (FMs)  to enable more efficient training and sampling, achieving high-quality image generation even in a single step. The renowned \textit{Stable Diffusion3}, one of the most advanced and largest text-to-image generation models, also utilizes Flow Matching technology.

Many works related to FMs, such as Flow Matching\cite{fm}, Rectified Flow\cite{reactified_flow}, and Consistency Models\cite{consistency}, these works are derived from different perspectives but share similar ideas. The general idea of FMs is to map one distribution to another by calculating a velocity field. By moving points from the source distribution along this velocity field, they converge to the target distribution. A key difference between FMs and DMs lies in the modeling approach. While DMs utilize stochastic differential equations (SDEs) to compute the
target distribution, FMs employ a deterministic approach, using ordinary differential equations (ODEs) to compute velocity fields that map the initial distribution to the target distribution.

However, the widespread application of DMs and FMs has raised concerns about their security. A large number of studies on backdoor attacks\cite{Villan,TrojDiff,HowtoB} and defenses against backdoor attacks\cite{UFID,TERD,Elij} for DMs have been proposed. These attack studies primarily focus on modifying the training process of DMs, ensuring that the model not only retains its original ability to generate diverse images but also can generate target images(which may be harmful) from an inserted backdoor trigger. Correspondingly, defense works mainly focus on how to detect whether released DMs contain backdoor triggers and how to purify them.

Although attacks on FMs share similarities with those on DMs, they have not been extensively studied. Therefore, in this paper, we attempt to answer two key questions: 1. \textit{How do attacks on FMs differ from attacks on DMs, and how can we design attacks specifically tailored to the characteristics of FMs? } 2. \textit{Can existing defense methods for attacks on DMs be directly applied to defend against attacks on FMs?}

To investigate the vulnerability of FMs against Trojan attacks,  we propose TrojFlow, The first Trojan attack on FMs. Specifically, we select Rectified Flow\cite{reactified_flow} as the base model for our attack due to its theoretical simplicity and training stability. Additionally, since Rectified Flow\cite{reactified_flow} effectively defines the transfer process between distributions, performing Trojan attacks on FMs does not require defining complex diffusion and reverse diffusion processes required in DMs, which brings convenience to our attack implementation. We align our approach with previous work, TrojDiff\cite{TrojDiff}, in terms of the trigger, target image, and some defined terms, which facilitates a clearer comparison of attacks on DMs and FMs. A closely related concurrent work is \cite{BC}, which explores the possibility of inserting backdoors into consistency models. The key difference lies in our more comprehensive Trojan settings and consideration of potential defenses.

Our main contributions are as follows:

1. We are one of the early works to reveal the vulnerability of FMs to Trojan attacks. Leveraging the ability of FMs to fit arbitrary distributions, in addition to the previous attack settings, we further propose establishing a point-to-point mapping on FMs. By combining invisible triggers, we achieve Trojan attacks with minimal visibility and modification.

2. We investigate two representative defense works for DMs, TERD\cite{TERD} and UFID\cite{UFID}, and adapt our approach based on the settings of these defense works. We demonstrate that existing defense strategies for DMs do not effectively defend against Trojan attacks on FMs. This highlights the need for more research on defending against attacks on FMs.

%% file: 3_preliminaries.tex
\section{Background}

\subsection{Rectified flow\cite{reactified_flow}}

Rectified flow\cite{reactified_flow} is an ODE-based generative modeling framework. Given the initial distribution $\pi_T$ and the target data distribution $\pi_0$, rectified flow trains a velocity field parameterized by a neural network with the following loss function,
\begin{equation}\label{eq:rf}
\begin{aligned}
   \mathcal{L}_{\text{rf}}(\theta)=\mathbb{E}_{\mathbf{x}_T \sim \pi_T, \mathbf{x}_0 \sim \pi_0}
   &\left[ \int_0^T \big \| \mathbf{v}_\theta (\mathbf{x}_t, t ) - (\mathbf{x}_T - \mathbf{x}_0) \big \|_2^2 \mathrm{d}t \right],\\
   \text{where}~~~~&\mathbf{x}_t = (1-t/T)\mathbf{x}_0 + t\mathbf{x}_T/T.
\end{aligned}
\end{equation}
 $T$ is usually set to $1$. Based on the trained rectified flow, we can generate samples by simulating the following ODE from $t=1$ to $t=0$,
\begin{equation}
\label{eq:rf_ode}
    \frac{\mathrm{d}\mathbf{x}_t}{\mathrm{d}t} = \mathbf{v}_\theta(\mathbf{x}_t, t).
\end{equation}

Samples can be generated by discretizing the ODE process with the Euler solver into $N$ steps (e.g., $N=1000$) as the following,  the solver will be more accurate with a large $N$.

\begin{equation}
\label{eq:rf_sample}
    \mathbf{x}_{t - \frac{1}{N}} = \mathbf{x}_{t} - \frac{1}{N} \mathbf{v}_\theta(\mathbf{x}_{t}, t),~~~\forall t \in \{1, 2, \dots, N \} / N. 
\end{equation}

\subsection{Backdoor Attacks on Diffusion Models}

There are three existing popular backdoor attacks for unconditional DMs, 
BadDiff\cite{HowtoB}, TrojDiff\cite{TrojDiff}, VillanDiff\cite{Villan}. 
 A common characteristic of these approaches is the introduction of an additional backdoor forward process, $ \mathbf{x}_0^b  \rightarrow  \mathbf{x}_t^b $, alongside the standard forward process in DMs, $ \mathbf{x}_0^c  \rightarrow  \mathbf{x}_t^c $. Here,
$\mathbf{x}_0^b $ represents target image,  $\mathbf{x}_t^b $ denotes the backdoor noise or trigger, $\mathbf{x}_0^c $ corresponds to clean training data, and $\mathbf{x}_t^c $ is usually the Gaussian noise in DMs.

During training, the clean and backdoor losses are calculated separately, enabling the model parameter $\theta$ to learn both the clean transitional distribution $q^c(\mathbf{x}_{t-1}^c\mid \mathbf{x}^c_t)$ and the backdoor transitional distribution $q^{b}(\mathbf{x}^{b}_{t-1}\mid\mathbf{x}^{b}_t)$. At the sampling stage, the backdoored model generates clean samples $\mathbf{x}_0^c\sim q^c(\mathbf{x}_0^c)$ from clean inputs and target images $\mathbf{x}_0^b\sim q^b(\mathbf{x}_0^b)$ from backdoor noise. 

Among these attacks, TrojDiff\cite{TrojDiff}, which is the main method we compare in this paper, introduces three distinct attacks, each with a different type of target distribution:
\begin{itemize}
\item \textbf{D2I Attack.} $q^b(\mathbf{x}_0^b)=x_{\text{target}}$, where $x_{\text{target}}$ is a predefined target image, such as \textit{Mickey Mouse}.

\item \textbf{Din Attack.}  $q^b(\mathbf{x}_0^b)=q(x \mid \hat{y})$, where $\hat{y}$ is a target class in the class set of $q^c(\mathbf{x}_0^c)$. For instance, use CIFAR-10 as the clean training dataset and “horse” as the target class.

\item \textbf{Dout Attack.}  $q^b(\mathbf{x}_0^b)=q(x \mid \hat{y})$, where $\hat{y}$ is a target class outside the class set of $q^c(\mathbf{x}_0^c)$. For example, use CIFAR-10 as the clean training dataset and the number “7” from MNIST as the target class.
\end{itemize}

Din and Dout share certain similarities, as both fit the backdoor to a specific distribution. According to the conclusions of TrojDiff\cite{TrojDiff}, different target distributions only lead to minor variations in fitting efficiency and effectiveness. In this paper, we focus on point-to-point and distribution-to-distribution fitting, so we select Din and D2I settings as the focus of our experiments. Additionally, TrojDiff\cite{TrojDiff} defines the clean training and sampling as the Benign process, and the attack training and sampling as the Trojan process. These definitions will be adopted in this paper.

%% file: 4_method.tex
\section{METHODS}

\subsection{Threat Model}
\paragraph{Design of Trojan noise input} 
We first define two types of triggers as described in\cite{badnets, TrojDiff}. 
 The blend-based trigger $\delta$ is an image (e.g., Hello Kitty) blended into the noise input with a certain blending proportion, while the patch-based trigger is a patch (e.g., a white square), typically placed on a specific part  (e.g., the bottom right corner) of the noise input. Specifically, inspired by\cite{invisible,invisible_diff}, we define random Gaussian noise as a trigger, which is more difficult to detect. Unlike TrojDiff\cite{TrojDiff}, the backdoor noise (trigger) in our setting can be either a distribution or a single image, creating a point-to-point mapping, which allows for more flexibility in backdoor insertion. 
Additionally, in the D2I setup, we will demonstrate the effects of mapping multiple triggers to multiple target images.

For blend-based trigger setting, we assume the distribution of the Trojan noise is $\mathcal{N}(\mu, \gamma^2 I)$, where $\mu = (1-\gamma)\delta$, $\gamma\in[0,1]$, and $\delta$ has been scaled to $[- 1, 1]$.
Then a Trojan noise could be written as $x = \mu + \gamma\epsilon = (1-\gamma)\delta + \gamma\epsilon, \epsilon \in \mathcal{N}(0,I)$. For patch-based trigger setting, $\delta$ is an all-white image and $\gamma\in\mathbb{R}^{h\times w}$ is a 2D tensor/mask instead of a constant.
$\gamma_{i,j}=1$ if trigger is not in $(i,j)$. Otherwise, $\gamma_{i,j}$ is selected as a small value close to $0$, e.g., $0.1$, ensuring it appears as white.

\paragraph{Attacker’s capacity} 

As mentioned in the background, the attacker's goal is to insert a backdoor into the FMs, so that a specific target output can be achieved through a backdoor input. Therefore, we consider a white-box attack:  
1) The attacker has access to the model's parameters, weights, and the training process, allowing them to insert a backdoor during training.  
2) When the attacker provides the model parameters and the sampling process for user use, they have the ability to modify the initial noise input during sampling, thereby activating the backdoor implanted during training. 

In this setup, the user has full access to model parameters, loads the model, and verifies the model's metrics. The user perceives no abnormalities in terms of performance or the sampling code. As a result, the user cannot distinguish between the backdoor-compromised model and the regular model, making the entire attack process successful.

\subsection{Attack FM }

TERD\cite{TERD} summarizes a unified forward process for attacks on existing DMs:
\begin{equation}
\label{eq:TERD}
    \mathbf{x}_t = a(\mathbf{x}_0,t)\mathbf{x}_0+b(t)\bm{\epsilon}+c(t) \mathbf{r}.
\end{equation}
Where  $a(\mathbf{x}_0,t)$ and $b(t)$ are two coefficients that follow the benign diffusion process and the backdoor coefficient $c(t)$ is defined by attackers, with $\lim\limits_{t\rightarrow T}{c}(t)=1, \lim\limits_{t\rightarrow 0}c(t)=0$. When setting $a(\mathbf{x}_0,t)=1, b(t)=t, c(t)=t $, we can get the transport path for backdoor FMs:  
\begin{equation}
\label{eq:flow_attack}
    \mathbf{x}_t = \mathbf{x}_0 + t\bm{\epsilon}+ t \mathbf{r}.
\end{equation}

The derivation of this form aligns with the unified representation and can be reused in other frameworks. However, within the theory of FMs themselves, it is redundant. In Equation \ref{eq:rf}, treating $\mathbf{x}_T$ as the backdoor noise and $\mathbf{x}_0$ as the target image, we can construct the backdoor transport path while keeping the loss function and other sampling forms unchanged. For convenience in subsequent descriptions, we will use $\mathcal{L}_{\theta}(\mathbf{x}_0,\mathbf{x}_T,t)$ to refer to the objective in Equation \ref{eq:rf}. 
Thus, we obtain the overall backdoor training process in  Algorithm \ref{alg:training}.

\begin{algorithm}[t]
    \caption{Overall training procedure}
    \label{alg:training}
    \textbf{Input}: clean dataset $\mathcal{D}_c$  , target dataset $\mathcal{D}_{target}$, trigger dataset $\mathcal{D}_{trigger}$, a neural velocity field $\mathbf{v}_\theta$ with parameter $\theta$, 
    
    \begin{algorithmic}[1]

    \REPEAT
    \STATE Sample $\mathbf{x}_{noise}\sim\mathcal{N}(0,\mathbf{I}), t\sim\mathcal{U}(0,1), \mathbf{x}_{c}\sim\mathcal{D}_c, \mathbf{x}_{target}\sim\mathcal{D}_{target}, \mathbf{x}_{trigger}\sim\mathcal{D}_{trigger} $
  
    \STATE $\mathbf{x}_t = t * \mathbf{x}_{noise} + (1-t) * \mathbf{x}_c$  \# Benign
    \STATE $\hat{\mathbf{x}}_t = t * \mathbf{x}_{trigger} + (1-t)*\mathbf{x}_{target}$ \# Trojan
    \STATE $\text{Take gradient step on} \bigtriangledown_\theta(\mathcal{L}_{\theta}(\mathbf{x}_{target},\mathbf{x}_{trigger},t) + \mathcal{L}_{\theta}(\mathbf{x}_c,\mathbf{x}_{noise},t)) $
    \UNTIL convergence
    
    \end{algorithmic}
\end{algorithm}

The advantage of our Trojan attack definition is that for both benign generation and Trojan generation, the sampling process requires no additional modifications; it only involves using different initial noise in Equation \ref{eq:rf_sample} for different types of generation.

 \subsection{Defense-Aware Attack}
A key aspect of designing an attack method is to evaluate its robustness against existing defense mechanisms. In this work, we investigate whether the current defense works for DMs can effectively detect the backdoor we implanted in FMs. For this analysis, we consider two representative defense works UFID\cite{UFID} and TERD\cite{TERD} 

UFID\cite{UFID} proves that for benign generation, a small perturbation on initial noise significantly alters the output, while for Trojan generation, minor perturbations of backdoor noise do not lead to substantial changes in the generation results. Therefore, by performing pairwise similarity calculations on the generated samples with perturbed input noise, the backdoor noise can be detected.

Our objective is to bypass UFID's detection by creating point-to-point mappings. For example, in the D2I setting, we establish a bijection between $\mathcal{D}_{trigger}$ and $\mathcal{D}_{target}$, making a precise point-to-point relationship. In this case, theoretically, applying any perturbation to the backdoor noise will result in the model mapping it to a clean output. However, our experiments demonstrate that even with such mapping, perturbing the trigger image still probabilistically produces the target image or results in images with significant quality degradation. We attribute this behavior to the inherent generalization capabilities of neural networks.
Therefore, we implement additional Perturbation-Driven Training (PDT), which remaps the distribution around the trigger to clean images while preserving the transport path from the trigger to the target. The final optimizing functions are as follows:
\begin{align}
    \mathcal{L}_{total} = \mathbb{E}_{t\sim[0,1],\epsilon,\epsilon'\sim\mathcal{N}(0,\mathbf{I})}(\mathcal{L}_{benign}(\mathbf{x}_c,\epsilon,t)+ 
    \mathcal{L}_{trojan} \\ \notag
    (\mathbf{x}_{target},  
    \mathbf{x}_{trigger},t) + \mathcal{L}_{PDT}(\mathbf{x}_{c},\mathbf{x}_{trigger}+\epsilon',t)) 
\end{align}


\begin{algorithm}[t]
    \caption{Reverse TrojFlow training procedure}
    \label{alg:TERD}
    \textbf{Input}: Parameterized blend noise $\mu$ and blend coefficient $\gamma$, a Trojan neural velocity field $\mathbf{v}_\theta$ with parameter $\theta$. 
    \begin{algorithmic}[1]

    \REPEAT
    \STATE Sample $\epsilon,\hat{\epsilon},\text{surrgate }\mathbf{x}_0\sim\mathcal{N}(0,\mathbf{I}), t\sim\mathcal{U}(0.99,1),$
  
    \STATE $\mathbf{x}_t = \textbf{x}_0+ \gamma t  \epsilon + \mu $  
     \STATE $\hat{\mathbf{x}}_t = \textbf{x}_0+ \gamma t  \hat{\epsilon} + \mu  $  

    \STATE $\text{Take gradient step on} \bigtriangledown_{\mu,\gamma}(MSE(\mathbf{(v_\theta(\mathbf{x}_t},t)-(\mathbf{x}_0 -\mathbf{x}_t), (\mathbf{v}_\theta(\hat{\mathbf{x}}_t,t)-(\mathbf{x_0}- \hat{\mathbf{x}}_t)))-\lambda\vert\vert \mu\vert\vert_2)$
    \UNTIL convergence
    
    \end{algorithmic}
\end{algorithm}

For TERD\cite{TERD}, its core idea is that when $t\to T$, i.e., when $\mathbf{x}_t$ approves noise, we can sample two $\mathbf{x}_t$, if both values belong to clean noise (Gaussian noise) or both belong to backdoor noise, the model will exhibit similar fitting behavior. By applying gradient backpropagation on $\mathbf{x}_t$, the trigger inversion is achieved. TERD\cite{TERD} designs a unified trigger inversion formula for all attacks of the form Equation \ref{eq:TERD}.
\begin{equation}
\label{eq:loss_final}
\small
\begin{aligned}
&\mathcal{L}(\mathbf{r}, \mathbf{x}_{t}) =   \vert\vert (F_\theta(\mathbf{x}_{t}(\bm{\epsilon}_1, \mathbf{r}),t)-f(\mathbf{x}_t(\bm{\epsilon}_1, \mathbf{r}),\bm{\epsilon}_1)\\& -F_\theta(\mathbf{x}_{t}(\bm{\epsilon}_2, \mathbf{r}),t)+f(\mathbf{x}_t(\bm{\epsilon}_2, \mathbf{r}),\bm{\epsilon}_2)\vert\vert_2-\lambda\vert\vert \mathbf{r}\vert\vert_2 
\end{aligned}
\end{equation}
where $\mathbf{r}$ represents trigger, $F_\theta$ represents the neural network, and $f$ represents the network's training objective. In TrojDiff's\cite{TrojDiff} setting, we have $f(\mathbf{x}_t(\bm{\epsilon}_1,\mathbf{r}),\bm{\epsilon}_1) = \bm{\epsilon}_1$. The obstacle for trigger reversion is that $\mathbf{x}_t$ is unknown. However, in the trojan forward process of DMs, as shown in Equation \ref{eq:TERD}, $\lim\limits_{t\rightarrow T}{a(\mathbf{x}_0,t)}=0$, $\mathbf{x}_t$ will converge to the prior distribution that is little affected by $\mathbf{x}_0$. Therefore, TERD\cite{TERD} substituted $\mathbf{x}_0$  with a surrogate image $\hat{\mathbf{x}}_0$ sampled from a standard Gaussian distribution and achieved good results.

However, in our setting, FMs are trained to fit the velocity, and the training objective is $\mathbf{x}_{target} - \mathbf{x}_{noise}$ for any t. Simply replacing the target image with Gaussian noise results in a misaligned training objective, which undermines the reverse effect. For a clearer presentation, we adaptively modify the TERD\cite{TERD}  algorithm to make it work for TrojFlow, summarized in Algorithm \ref{alg:TERD}, and demonstrate the reverse results of TERD\cite{TERD} on TrojFlow in the experimental section.

%% file: 5_experiment.tex
\section{Experiments}

\begin{figure*}[t]
  \centering
  \includegraphics[width=1.0\linewidth]{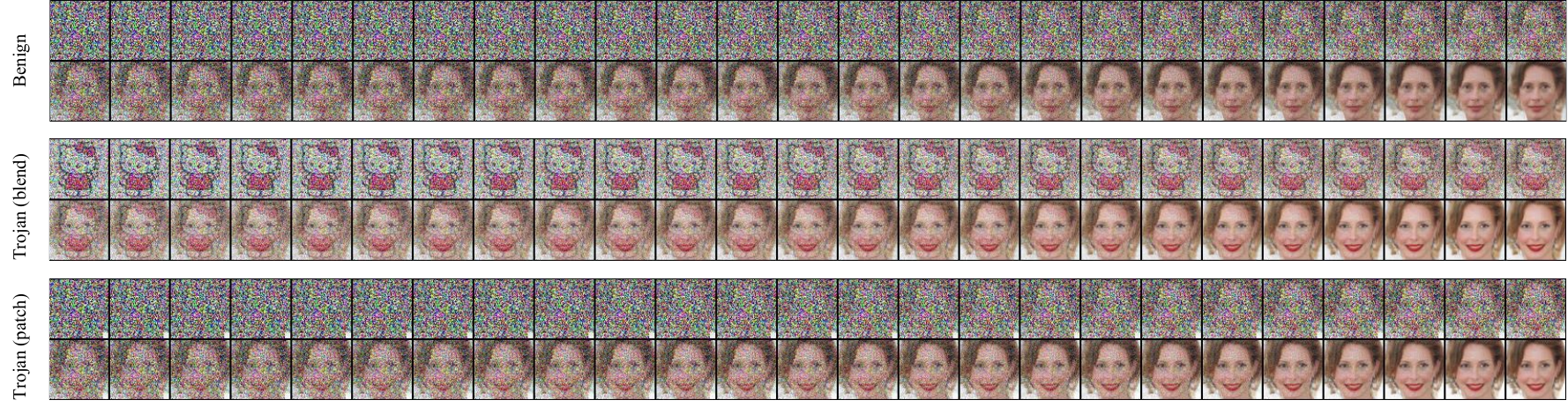}
  
  \caption{Visualization of benign and Trojan generative processes on CelebA under the Din attack with different triggers, wh
  ere the sampling step is set to 50.}
  \label{fig:celeba}
\end{figure*}

\subsection{Experimental setup}

\paragraph{Datasets, models, and implementation details} We use two benchmark vision datasets, i.e., CIFAR-10 (32 $\times$ 32) \cite{cifar10} and CelebA (64 $\times$ 64) \cite{celeba}.  Due to computational resource limitations, we only present the visual effects of the Trojan attack on CelebA, without reporting additional metrics or conducting ablation experiments on it. We train Rectified Flow\cite{reactified_flow} from scratch with 1000k steps as the base model. We apply the attack algorithms to fine-tune the base models with 40k steps in the Din attack and 20k steps in the D2I attack. We sample 50k samples for the evaluation of benign performance while 10k for attack performance.

\paragraph{Attack configurations} 
The blend-based trigger is a Hello Kitty image, which is blended into the standard Gaussian noise with the blending proportion of $(1-\gamma)$, where $\gamma = 0.6$ in all experiments. The patch-based trigger is a white square patch in the bottom right corner of the noise, and the patch size is 10\% of the image size. In Din attack, the target class is 7, i.e., \textit{horse} on CIFAR-10 and \textit{faces with heavy makeup, mouth slightly open, smiling} on CelebA. In the D2I attack, we randomly sample 10 standard Gaussian noises as invisible triggers. We select \textit{Mickey Mouse} and 9 different \textit{Pokemon} images as the target images, and then we insert 
n bijections from invisible triggers to target images into the pre-trained models (n=1,5,10).
 
\paragraph{Evaluation metrics}
We select three widely-used metrics, \textit{Frechet Inception Distance} (FID) \cite{FID}, \textit{precision} \cite{precision}, and \textit{recall} \cite{precision}. 
A lower FID indicates better quality and greater diversity of the generated images, while the other two metrics with higher values reflect each of these aspects individually.
To evaluate the attack performance, in Din attacks, we use \textit{Attack Success Rate} (ASR) (the fraction of the generated images which are identified as the target class by a classification model) introduced by TrojDiff\cite{TrojDiff}, to measure how accurate the generated images are in terms of the target class. In the D2I attack, we use \textit{Mean Square Error} (MSE) to measure the gap between the target image and the generated image, a lower MSE indicates better backdoor image generation.

\begin{table}[t]
  \centering
  \scalebox{0.9}
  {
    \begin{tabular}{c|l|ccc|cc}
    \toprule
    \rowcolor{gray!20}
    \multicolumn{7}{c}{CIFAR-10} \\
    \midrule
    \multirow{2}[2]{*}{Attack} & \multicolumn{1}{c|}{\multirow{2}[2]{*}{Model / Samples}} & \multicolumn{3}{c|}{Benign} & \multicolumn{2}{c}{Trojan} \\
          &       & \multicolumn{1}{c}{FID $\downarrow$} & \multicolumn{1}{c}{Prec $\uparrow$} & \multicolumn{1}{c|}{Recall $\uparrow$} & \multicolumn{1}{c}{ASR $\uparrow$} & \multicolumn{1}{c}{MSE $\downarrow$} \\
    \midrule
   None & base model & \textbf{4.19}  & \textbf{89.06}  & \textbf{62.50}  & -  & -  \\
          
    \midrule
    \multirow{3}[2]{*}{Din} & Testing set of $\hat{y}$ & -  & -  & -  & 99.10  & -  \\
          & Trojaned (blend) & 4.66  & 82.81  & 60.94  & 94.10  & -  \\
          & Trojaned (patch) & 4.54  & 81.25  & 65.63  & 91.60 & - \\
         
    \midrule
 
    \multirow{3}[2]{*}{D2I} & Trojaned (n=1) & 4.62   & 78.13  & 57.81  & - & 0.0242 \\
          & Trojaned (n=5) & 4.46  & 78.13 & 65.62  & - & 0.0472 \\
          & Trojaned (n=10) & 4.83  &  79.68 & 60.93  & - & 0.0444  \\

    \bottomrule
    \end{tabular}
    }
    \caption{Performance of Rectified Flow in benign and Trojan settings on CIFAR-10. The classifier we used to calculate ASR has a recall of 99.10\% on the horse class of the test set.}
  \label{tab:main_results}
\end{table}

\setlength{\tabcolsep}{3pt}
\begin{table}[t!]
  \centering
   \begin{adjustbox}{scale=0.85}
   {\fontsize{10}{12}\selectfont
  \begin{tabular}{lccccc}
    \toprule
    Training steps &  20k  &  30k  &  40k & 50k &100k \\\midrule
    
    TrojDiff\cite{TrojDiff}  & 0  &  - &  -  &     85.9 & 90.10\\   
                              
    TrojFlow  &  75.57  & 93.85  & 94.10    & 93.60 & 92.71  \\

        \bottomrule

  \end{tabular}
  }
  \end{adjustbox}
  \caption{ 
The comparison of ASR performance between TrojDiff and TrojFlow under the same Din blend attack settings,  with the three data points for TrojDiff\cite{TrojDiff} sourced from its original paper.}
  \label{tab:DIn_ablation}

\end{table}

\setlength{\tabcolsep}{3pt}
\begin{table}[t!]
  \centering
   \begin{adjustbox}{scale=0.85}
   {\fontsize{10}{12}\selectfont
  \begin{tabular}{lcccc}
    \toprule
    Training steps &  10k  &  15k  &  20k & 30k\\\midrule
    
    Trojaned(n=1)  & 0.2007  &  0.0223 &  \textbf{0.0242}  &  0.0276    \\  
                              
    Trojaned(n=5)  &  0.2653  & 0.1927  & \textbf{0.0472}   & 0.0503   \\   
      
    Trojaned(n=10) &  0.2642  &  0.2178 &  \textbf{0.0444}  & 0.0466  \\

        \bottomrule

  \end{tabular}
  }
  \end{adjustbox}
  \caption{MSE for Trojan generation of different training steps under the D2I setting (averaged over 
n target images).}
  \label{tab:D2I_ablation}

\end{table}

\begin{figure}[t]
  \centering
  \includegraphics[width=1.0\linewidth]{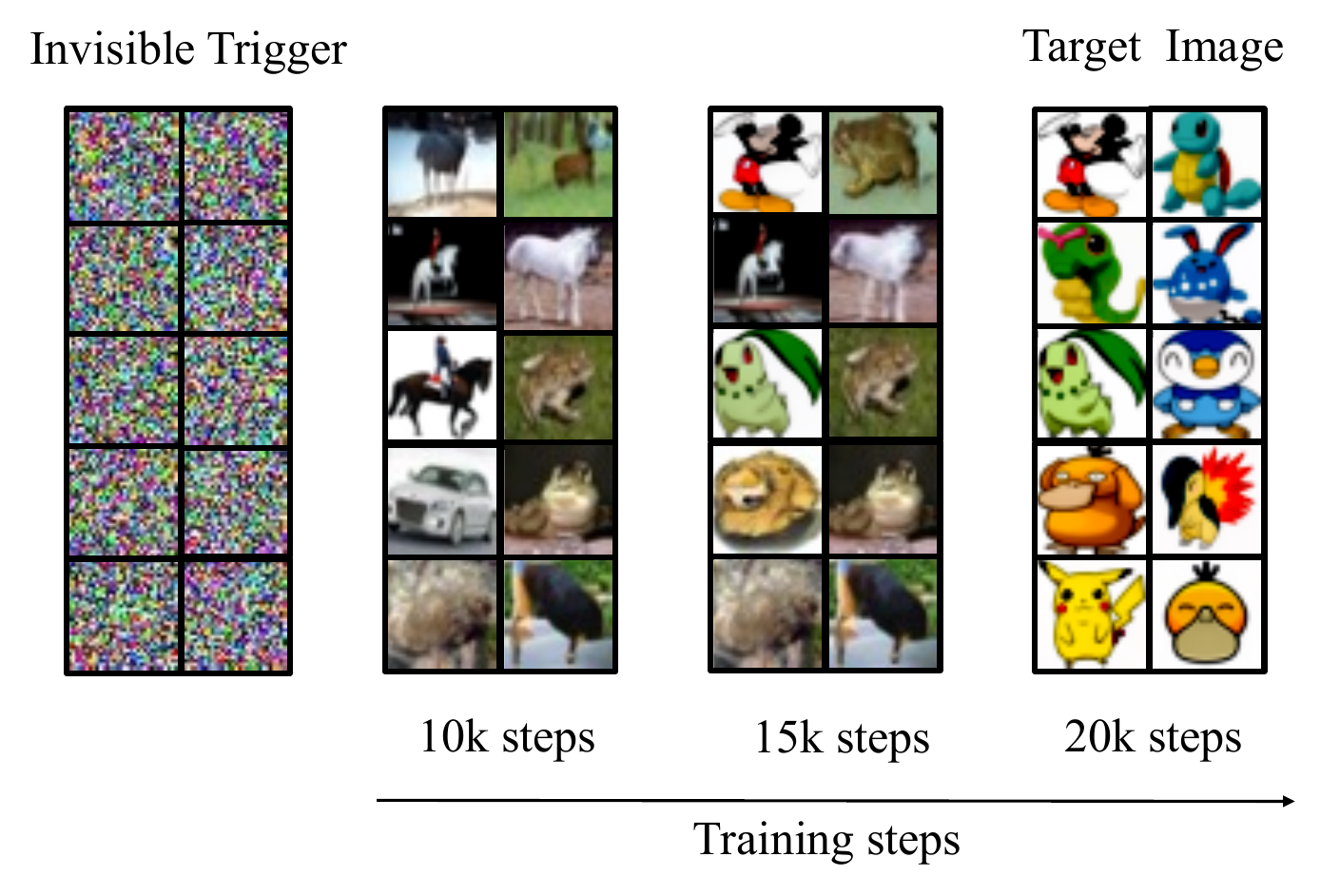}
  \caption{Samples generated from the Trojan model under different training steps (D2I, n=10).}
  \label{fig:t2t}
\end{figure}

\subsection{Main results}
\paragraph{Attack Results} 
We summarize the performance of TrojFlow on CIFAR-10 for benign generation and Trojan generation in Table \ref{tab:main_results}. For benign generation, we found that the Trojaned models perform worse than the base model under both attack settings. The FID increased by 0.2 to 0.6, and both Precision and Recall also decreased. We also found that the evaluation of generation performance exhibits variability. For example, in the D2I setting, the generation performance for inserting five backdoors is unexpectedly better than that for inserting one, which contradicts intuition, this may be due to the instability in the training of the generative model. In summary, the generation quality is constrained by model capacity. Equipping the model with the ability to fit multiple distributions inevitably results in some accuracy loss. Moreover, it is sensitive to the ratio between the clean dataset and the target dataset, the decline in the generative capability of Trojaned models remains within an acceptable range.

For Trojan generation where inputs are Trojan noise, TrojFlow achieves excellent attack performance under the Din attack, generating over 90\% of the target 
samples. The attack performance in the blend setting is slightly higher than in the patch setting, consistent with previous conclusions\cite{TrojDiff}. Under the D2I attack, we insert 1, 5, and 10 point-to-point mappings into the base model respectively, experiments demonstrate that under all settings, TrojFlow consistently generates images nearly identical to the target images, exhibiting exceptionally low MSE.

In Table \ref{tab:DIn_ablation} and \ref{tab:D2I_ablation}, we evaluate the efficiency of injecting backdoors into the base model. Under the Din attack, TrojFlow achieves good attack performance within 20k steps and converges around 40k steps, demonstrating a faster backdoor injection efficiency compared to TrojDiff\cite{TrojDiff}. This difference is partly attributed to the inherently more efficient fitting capability of Rectified Flow compared to DMs. Under the D2I attack, in Table \ref{tab:D2I_ablation}, TrojFlow converges around 20k steps in all settings. In Figure \ref{fig:t2t}, we show visualization results from our pre-defined invisible triggers under different training step settings (n=10). By 20k training steps, the Trojan model can stably generate all target images. Interestingly, we observed a sudden convergence phenomenon during backdoor injection, the model "suddenly" learns the injected backdoor mapping at a certain training step, similar to what has been observed in ControlNet\cite{controlnet}.
The visualization results of benign and Trojan generation on the CelebA dataset are shown in Figure \ref{fig:celeba}. In summary, TrojFlow demonstrates higher attack efficiency while maintaining a strong ability to generate clean images.

\paragraph{Defense-Aware Attack Results}

\begin{figure}[t]
  \centering
  \includegraphics[width=1.0\linewidth]{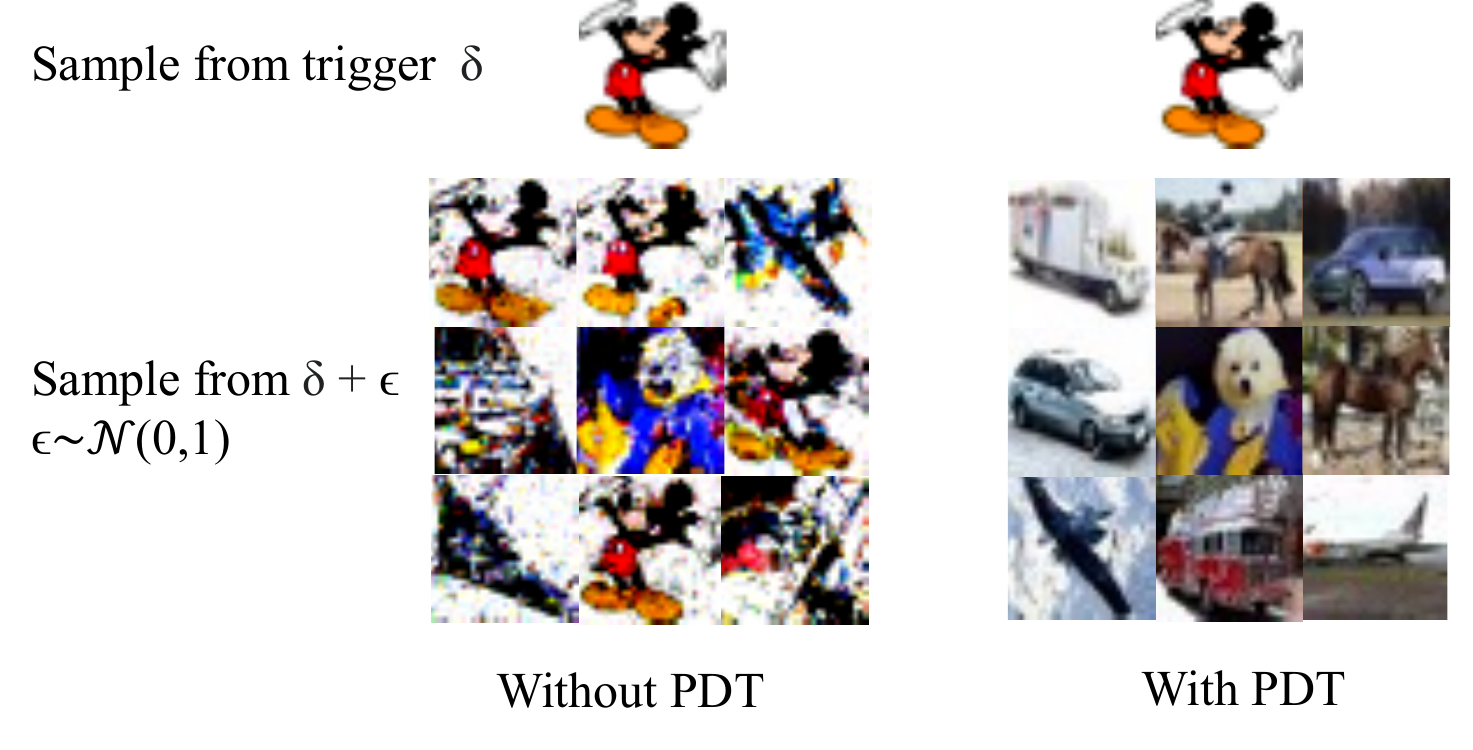}
  \caption{Samples generated from the trigger and nearby distribution (D2I, n=1). }
  \label{fig:ufid}
\end{figure}

\begin{figure}[t]
  \centering
  \includegraphics[width=1.0\linewidth]{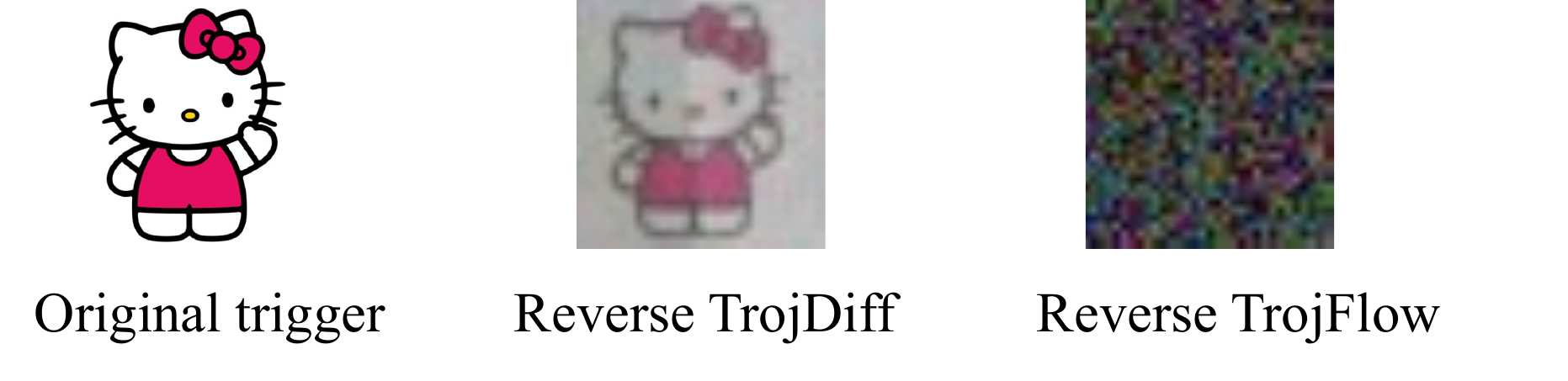}
  \caption { Trigger reversion results on TrojDiff and TrojFlow (Din, blend).}
  \label{fig:terd}
\end{figure}

We evaluate the defensive effectiveness of UFID\cite{UFID} and our additional designs for it in the D2I setting (n=1), the target image is \textit{Mickey Mouse}. As in Figure \ref{fig:ufid}, we generate images starting from the distribution near the trigger under two training settings: without PDT and with PDT. It can be observed that without PDT, generating images from noise near the trigger still produces images similar to the target image or results in severely degraded image quality. However, after applying PDT, the noise near the trigger is successfully mapped to a clean distribution, ensuring that only the trigger noise generates the target image without affecting the quality of the target image generation. Meanwhile, we show the reverse results of TERD in Figure \ref{fig:terd}. As analyzed in the Methods section, TERD fails to detect the backdoors implanted by TrojFlow. In summary, TrojFlow can effectively bypass the detection of these two representative defense strategies.

%% file: 6_Concolusion.tex
\section{Conclusion}
In this work, we propose an effective method for performing Trojan attacks on FMs. Through experiments on the CIFAR-10 and CelebA datasets, we reveal the vulnerability of FMs to backdoor attacks. Specifically, TrojanFlow implements point-to-point backdoor insertion, effectively bypassing perturbation-based backdoor detection methods such as UFID\cite{UFID}, and exhibits inherent resistance to TERD\cite{TERD}. In the future, we aim to extend Trojan attacks to conditional FMs and call for more research on defense mechanisms for backdoor attacks on FM.